# Unjustified Sample Sizes and Generalizations in Explainable AI Research: Principles for More Inclusive User Studies

Uwe Peters
(1) Leverhulme Centre for the Future of Intelligence
University of Cambridge, UK
(2) Center for Science and Thought, University of Bonn, Germany
Mary Carman
Department of Philosophy
University of the Witwatersrand, South Africa

*Abstract* – Many ethical frameworks require artificial intelligence (AI) systems to be explainable. Explainable AI (XAI) models are frequently tested for their adequacy in user studies. Since different people may have different explanatory needs, it is important that participant samples in user studies are large enough to represent the target population to enable generalizations. However, it is unclear to what extent XAI researchers reflect on and justify their sample sizes or avoid broad generalizations across people. We analyzed XAI user studies ($n = 220$) published between 2012 and 2022. Most studies did not offer rationales for their sample sizes. Moreover, most papers generalized their conclusions beyond their study population, and there was no evidence that broader conclusions in quantitative studies were correlated with larger samples. These methodological problems can impede evaluations of whether XAI systems implement the explainability called for in ethical frameworks. We outline principles for more inclusive XAI user studies.

## Introduction

AI systems used for decision-making in high-stakes contexts (e.g., hiring domains or medical predictions) are often computationally opaque, i.e., they follow complex algorithms that even AI engineers can no longer fully understand [1]. Since this makes the trustworthiness of these systems questionable, many ethical AI frameworks require the outputs of AI technology to be explainable to people so that they can potentially object to AI-based decision-making [2]. One main approach to implementing explainability in practice involves supplementing opaque AI with explainable AI (XAI) models designed to make opaque systems' outputs understandable to humans.

To evaluate whether XAI models' explanations satisfy a principle of explainability, developers often test their systems on human users [3]. Since explanation needs may vary across individuals, the value of such user studies commonly depends on whether these studies' results provide insights into human users' requirements that are generalizable from the particular study participants to broader populations [1]. If findings about XAI systems only hold for a small group of individuals, these systems may not meet explanatory requirements of other people affected by or using them and hence fail to adequately implement explainability in practice. This can impact trust in AI systems. The generalizability of XAI user study results is thus vital for developers and policy-makers to gain insight into whether a given XAI system satisfies an ethical principle of explainability.

One way of achieving generalizability in an XAI user study is to test the whole target population. However, researchers' constraints (e.g., funding) make this rarely feasible. Usually, in human-subject studies in general, a sample is taken from the target population and results are then extrapolated to that population. This can go wrong when the sample is not large enough to reflect the target population [4]. Thinking about and justifying a chosen sample size is thus important for generalizability.

Concerns about underpowered sample sizes, i.e., samples too small to detect effects and generalize results, have been raised across the sciences [5]. And many social science journals now require authors to provide sample size justifications, i.e., rationales for selecting a particular sample size (e.g., power analyses) [6]. However, while the analysis of these justifications is a topic of interest in different fields [6, 7, 8], no examination of sample size justifications in XAI user research yet exists.

To be sure, if XAI user researchers tailored their generalizations to their samples' size and composition such that the researchers only make broader claims when they have tested larger, more representative samples [9], then small or unrepresentative samples and omissions of sample size justifications may not be problematic. Similarly, if a given XAI user study was only run to quickly receive feedback and determine whether at least some users understand a particular XAI output then little generalizability would be sought and no sample size justification may be needed. However, no systematic analysis of how broadly or narrowly XAI researchers do in fact generalize from their studies exists so far.

Yet, investigating concerns about sample sizes, sample size justification, and generalizability in XAI user studies is especially important and intertwined with AI ethics. If XAI user studies' have shortcomings in sampling or generalizability, stake-holders may need to re-calibrate their trust in the XAI models that these studies may advertise because the studies' findings might then only apply to particular groups of individuals. The implementation of ethical frameworks for XAI systems thus also involves the implementation of certain basic principles of scientific methodology in user studies to ensure that people's trust in XAI systems is aligned with the systems' capability of providing widely acceptable explanations of opaque models.

We therefore systematically reviewed a large number of XAI user study papers ($n = 220$) published between January 2012 and July 2022 to examine sample size justifications and generalizations in them. Most of these studies did not offer sample size justifications, leaving it unclear whether their samples were large and representative enough for broad generalizations of results beyond the participants. However, most studies nonetheless generalized their results far beyond their samples. There was also no evidence that broader conclusions were correlated with larger samples, and such conclusions appeared even when researchers had not checked for relevant background knowledge (e.g., AI expertise) among their study participants that can affect result generalizability. These points suggest that *overgeneralizations*, i.e., conclusions whose scope is broader than warranted by the evidence and justification provided by the researchers, are pervasive in many available XAI user studies.

In summary, this paper offers the following novel contributions:

(1) It is the first systematic investigation of sample size justifications in XAI user studies that provides evidence that such justifications are rare in many available user studies.

(2) It offers the first quantitative data on overgeneralizations of results in XAI user studies.

(3) It introduces principles to mitigate these problems and be able to better evaluate whether XAI systems adequately implement explainability for ethical AI in practice.

## Background and related work

*Explainable AI.* While the literature on XAI is extensive, XAI methods can be broadly distinguished into transparent models and post-hoc (model-specific or model-agnostic) techniques [10]. Transparent models are by themselves understandable (e.g., logistic or linear regression) [11]. In contrast, post-hoc techniques aim to provide understandable information about how an opaque system produces its outputs for a given input, for instance, by accessing the system's internals (e.g., decision weights), or by analyzing the dependencies between input-output pairs to infer factors contributing to the system's decisions [10]. Post-hoc XAI processing can be local, aiming to explain specific outputs of opaque models, or global, aiming to explain the entire model's behavior, and XAI explanations may be visual (e.g., saliency maps), numerical (e.g., importance scores), or textual (e.g., feature reports) [12].

*Sample size justification.* If XAI user studies find that participants understand a particular XAI system's outputs well, this may be cited to claim that the system satisfies a key component of ethical frameworks for promoting trust in AI. This can facilitate the system's larger-scale deployment. Stakeholders thus need to have confidence that the conclusions drawn from XAI user studies are correct.

Confidence about study results relates to sample size in that narrow confidence intervals (CI), which indicate more precise results, require larger samples [9]. Larger samples also help detect small group differences for which smaller samples may not be sufficiently powered.

That said, even minute statistical effects can be boosted to significance by increasing sample size, as this reduces a statistic's standard error, meaning that one can find statistically significant but practically meaningless effects in very large samples. In that sense, samples can also be 'too big' for hypothesis testing and potentially magnify biases associated with mistakes in sampling or study design [13].

The sample size that a quantitative study needs depends on different factors that researchers may treat differently given their goals. These factors include the α-error level (capturing the risk of false positives, usually set at 5%), the β-error level (capturing the risk of false negatives, commonly set with a power of 80%), and the effect size [8]. Researchers also need to consider sample composition: While randomly chosen larger samples are more representative of a target population, large samples may be homogenous in relevant characteristics and so unrepresentative.

Other considerations to factor in are that, unlike large sample studies, small sample studies are cheaper and quicker to perform, which helps prevent wasting resources. And in qualitative research, having fewer participants can facilitate close associations with them, potentially providing more in-depth insights into personal experiences [14]. Since different decisions may be made based on these and other methodological factors, the provision of sample size justifications in XAI user study matters, as reviewers or other stakeholders cannot assume that the chosen sample is adequate for a given study [8].

Common approaches to justifying sample sizes in the social sciences include (1) power analysis, (2) heuristics (e.g., sampling guidelines, consistency with existing research, or rules of thumb), (3) pragmatic considerations (e.g., funding, participant availability, low response rate), and, for qualitative studies, (4) saturation, i.e., the point at which no new data (ideas, opinions, etc.) can be attained with more participants [4, 7, 8]. However, it is unclear whether any sample size justifications are reported in XAI user studies and how XAI researchers' generalizations are related to their sample sizes. A systematic review of XAI user studies is needed to investigate these issues.

## A systematic literature review

Following a rigorous methodology (as set out by [15]), we reviewed XAI user study papers to answer five research questions (RQ):

*RQ1.* Do researchers that conduct XAI user studies justify their sample sizes?

*RQ2.* Do XAI user studies with sample size justification have different (e.g., larger) sample sizes than those without?

*RQ3.* Do researchers that conduct XAI user studies restrict their conclusions to their participants or study populations or extrapolate beyond them?

*RQ4.* Are broader conclusions correlated with larger samples?

*RQ5.* Do researchers that conduct XAI user studies with, for instance, lay-users check whether their participants have a background (e.g., AI expertise) that can influence study results and generalizability?

**Methodology**

*Literature search.* We used three major databases: Scopus, Web of Science, and arXiv. Scopus and Web of Science index key computer science resources (e.g., ACM Digital Library, IEEExplore). ArXiv contains AI papers not yet published in journals, potentially providing insights into recent developments. We also used forward snowballing, i.e., we identified and included relevant papers frequently cited (>10 citations) in the studies selected for full-text analysis. The three databases were searched in July 2022 using search strings containing 15 variants of key words related to XAI ("XAI", "explainable AI", etc.) and end users ("user study", "user survey", etc., for details, see the Appendix). The results were 2523 papers. After removing duplicates ($n = 535$), titles and abstracts of the remaining 1988 papers were scanned for studies that met our selection criteria.

*Selection criteria.* We included any primary study (article, conference paper, book chapter) that surveyed people on their perception of AI-based explanations of automated processing and was published between January 2012 and July 2022. We excluded editorials, notes, opinion papers, reviews, or studies that did not contain information about sample size, were not in English, offered only a preliminary data analysis, or had a sample ≤ 5 (an exceedingly small sample to ensure study validity) [7]. 206 articles remained for further screening, during which forward snowballing produced an additional 14 papers, yielding a final $n = 220$ for full-text analysis (see PRISMA flowchart in the Appendix (Figure 1)).

During full-text analysis, we (two researchers) separately classified each paper using pre-specified criteria for extracting the following five sets of information.

(1) *General information.* We extracted publication year, study sample size (final participant number), and study design, categorizing papers as 'quantitative' (studies involving statistical analyses), 'qualitative' (studies involving qualitative analysis of interviews, free responses, etc.), or 'mixed' designs (studies involving both quantitative and qualitative elements).

(2) *Scope of conclusions.* In their articles, researchers may limit their study conclusions to their sample, a specific target population, or a minority of individuals by using the relevant qualifiers (e.g., 'our participants […]', 'U.S. users […]', 'many lay-people […]'). They may also use past tense to describe findings, limiting their conclusions to their study, sample, or study population. Papers with these features were classified as 'restricted'. Alternatively, authors may in their paper's abstract, results, discussion, or conclusion sections refer to users, experts, people, humans, etc. in general, not subsets of them (e.g., particular national target populations), or otherwise describe results in ways that suggest they hold across contexts, time, cultures, backgrounds, or groups. Papers with at least one such broad result-related claim in these sections were classified as 'unrestricted' (for examples, see Table 1). This label was also applied when an article additionally contained restricted claims, as papers usually undergo many revisions when authors can qualify their broader claims. If that does not happen, this suggests the authors consider their broader generalizations warranted.

(3) *User type.* Our categories were (a) 'lay-users' (for unspecified crowd-sourced participants,

unspecified students, or when the study material suggested a lay-user audience), (b) 'technical experts' (construed broadly as people with programming, IT, machine learning (ML), or computer science background), (c) 'lay-users/technical experts' (for (a) and (b)), (d) 'domain experts' (clinicians, lawyers, etc.), and (e) 'domain experts/technical experts' (for (d) and (b)).

(4) *Relevant background*. Many demographic factors (gender, age, etc.) may confound user study results. But including or excluding them as control variables requires justification because including control variables decreases available degrees of freedom, power, and may in some cases reduce the explainable variance available in outcomes of interest, facilitating false negatives, whereas excluding them can inflate the explainable variance in the criterion, facilitating false positives [16]. One factor that studies repeatedly found to produce different responses to XAI models is having a background in AI, ML, XAI, IT, or computer science [1, 17, 18]. Individuals with such backgrounds will display technical affinity and may already be familiar with salience maps, importance scores, etc. Asking XAI study participants about their potential technical background (i.e., technical affinity, expertise in programming, ML, XAI, IT, or computer science) to control for it during data collection or analysis is thus important if results are to be generalizable to lay-users. To record when researchers questioned or categorized users accordingly, we used a simple 'yes'/'no' classification per paper.

(5) *Sample size justification*. We operationalized sample size justification as any research effort before data collection to obtain adequate sample sizes. We coded papers ('yes'/'no') depending on the presence of at least one of the following: (a) power analysis, (b) heuristics (e.g., consistency with other studies), (c) pragmatic considerations (e.g., resource constraints), or (d) saturation.

*Reliability*. For each classification, inter-rater agreement was calculated (Cohen's κ) and was consistently above substantial (between κ = .71 and .90, $p < .001$). As a further reliability control, for the 'scope of conclusion' variable (our most complex classification), we additionally asked two independent, naïve raters to apply our pre-specified criteria to 25% of the data. Inter-rater agreement between their and our classifications was measured and was substantial (κ = .66 and .74, respectively). All remaining disagreements were resolved by discussion before the data were statistically analyzed ($α = .05$). Our materials are accessible on an OSF platform (https://osf.io/vzndw/).

**Results and discussion**

Most XAI user studies (97.7%, $n = 215$) were published (online or in journals) between 2018-2022, with an almost 50% increase from 2020 ($n = 46$) to 2021 ($n = 84$). Table I in the Appendix presents publication numbers by year. 57.3% ($n = 126$) of the reviewed studies used quantitative methods, followed by studies with mixed designs (30.4%, $n = 67$), and qualitative research (12.3%, $n = 27$).

*RQ1. Do researchers that conduct XAI user studies justify their sample sizes?* We found that 88.2% ($n = 194$) of the reviewed user study papers did not justify their sample sizes. Looking more specifically at papers by method type, from those with qualitative studies ($n = 27$), 0 offered a sample size justification. This is problematic because even though qualitative studies often do not aim for generalizability [14], they still require sample size justification, for instance, to ensure saturation [7]. Focusing on quantitative studies, for these studies, a power analysis is commonly viewed as the gold standard for evaluating study feasibility and for justifying sample size [4]. A power analysis is recommended to prevent under-sampling (i.e., too small, unrepresentative samples that can facilitate false negatives) and over-sampling (i.e., too large samples that can boost practically irrelevant effects to significance). We thus looked more closely at this particular kind of sample size justification in quantitative studies, setting aside qualitative and mixed study papers.

We found that of the 126 quantitative studies, 88.1% ($n = 111$) did not report any power analysis. Researchers who do not report such an analysis may still have conducted one. However, since conducting a power analysis is a methodological strength, it is hard to see why these researchers did not mention it in their papers if they had performed such an analysis. This suggests that such analyses and sample size justifications were not part of the study designs. For preliminary, piloting, or explorative studies, asking for a power analysis may be excessive. However, only 5% ($n = 11$) of the reviewed XAI studies indicated that their research was of that kind. Focusing on the more than 88% of the quantitative studies that did not provide any sample size justification, in not offering such a justification, their authors failed to establish that their selected samples were large enough to detect effects and generalize the study results beyond the participants.

*RQ2. Do XAI user studies with sample size justification have different (e.g., larger) sample sizes than those without?* It might be that sample size justification does not affect studies' sample sizes. If so, the sample sizes of studies with such justifications should not significantly differ from those without it. To examine this, we first tested our data for normality, finding that a non-parametric statistic was required ($p < .001$). Treating sample size justification as a binary variable and sample size (final participant $n$) as a scale variable, a Mann-Whitney $U$ test was run, showing that papers with sample size justification ($n = 26$) had significantly larger sample sizes (*Mean rank* = 141.23) than papers without it ($n = 194$) (*Mean rank* = 106.38, $U = 1723.00$, $z = -2.622$ $p = .009$). A rank-biserial correlation test additionally showed that sample size justification was correlated with larger samples, $r_{rb}$ (218) = .177, 95% CI [.042, .306], $p = .008$. To the extent that larger samples are more representative and increase generalizability, this finding suggests a link between sample size justification and generalizability-strengthening sample sizes. But how broadly exactly did XAI user study researchers generalize their results?

*RQ3. Do the researchers that conduct XAI user studies restrict their conclusions to their participants or study populations or generalize beyond them?* 68.2% ($n = 150$) of the 220 papers contained unrestricted conclusions, i.e., conclusions that referred very broadly to people, users, or humans as whole categories, or applied across context and time. Table 1 presents seven examples. Importantly, of the 150 'unrestricted' papers, 84.7% ($n = 127$) did not contain any sample size justification. Hence, even though their authors did not reflect on whether their sample sizes would support generalizations beyond the study population and so did not provide the needed epistemic basis for such generalizations, these papers nonetheless contained conclusions that extended vastly beyond the studied populations. Their authors thus overgeneralized their XAI user study results.

| |
|---|
| (1) "Specifically, our results suggest that users prefer more diverse local explanations when they are presented alone compared to when a global explanation is also available." [12] |
| (2) "We demonstrated that CX-ToM significantly outperforms baselines in improving human understanding of the underlying classification model." [26] |
| (3) "Our pilot study revealed that users are more interested in solutions to errors than they are in just why the error happened." [42] |
| (4) "Explanations lead people to view errors as being 'less incorrect', but they do not improve trust." [130] |
| (5) "People prefer item-centric but not user-centric or socio-centric explanations." [142] |
| (6) "Results indicate that human users tend to favor explanations about policy rather than about single actions." [205] |
| (7) "Our findings suggest that people do not fully trust algorithms for various reasons, even when they have a better idea of how the algorithm works." [213] |

**Table 1.** The number in square brackets refers to the identifier of the paper on the data spreadsheet from which the examples are taken (see https://osf.io/vzndw/).

*RQ4. Are broader conclusions in XAI user studies correlated with larger samples?* When considering quantitative studies, one may predict a correlation between broader claims and larger samples because randomly selected larger samples are more representative and so can ensure wider generalizability [9]. Finding such a link in *qualitative* studies is less likely, as these studies are often not intended to achieve broad result generalizability [14]. Since qualitative and mixed method studies (due to their qualitative component) may thus skew correlation tests, we excluded them from the analysis for *RQ4* and focused only on quantitative studies ($n = 126$). As data normality was violated ($p < .001$), we first conducted a Mann-Whitney *U* test to compare sample sizes between 'restricted' and 'unrestricted' papers. We found that 'unrestricted' papers ($n = 99$, *Mean rank* = 66.68) did not have statistically significantly larger samples than 'restricted' ones ($n = 27$, *Mean rank* = 51.83; $U = 1021.50$, $z = –1.873$, $p = .061$). A subsequent rank-biserial correlation test also did not provide evidence of an association between 'unrestricted' papers and sample size, $r_{rb}$ (124) = .168, 95% CI [–.013, .337], $p = .061$. Since 78.6% ($n = 99$) of the quantitative study papers were 'unrestricted', this suggests that many of the broad extrapolations that these papers contained were insufficiently supported: If they all had been sufficiently supported, we should have found unrestricted conclusions to be correlated with larger samples. However, it might still be that the papers' authors at least took care to control for potential confounding features in their study participants that may affect result generalizability.

*RQ5. Do researchers that conduct XAI user studies with, for instance, lay-users check whether their participants have a background (e.g., AI expertise) that can influence study results and generalizability?* Several studies had different kinds of users as their target audience. 69.1% ($n = 152$) had only lay-users, 10.5% ($n = 23$) only technical experts, 8.6% ($n = 19$) lay-users/technical experts, 6.4% ($n = 14$) only domain experts, and 4.1% ($n = 9$) domain experts/technical experts as intended targets. Importantly, focusing on the 152 studies that authors presented as testing only lay-users, 70.4% ($n = 107$) did not contain evidence that the participants (typically recruited via crowdsourcing platforms, e.g., MTurk) were questioned about or categorized according to their technical affinity, potential experience with XAI/ML, or computer science background. These participants may thus have had such a background. Since this can significantly influence people's perceptions of XAI outputs the studies' results cannot be broadly generalized [1, 17, 18]. Yet, we found that 78.5% ($n = 84$) of the studies did just that (they were 'unrestricted' papers).

## 4. Limitations

Having highlighted overgeneralizations in XAI research, our own study has several generality constraints. The studies we included were heterogeneous in user types, XAI outputs, sample sizes, and methods, making extrapolations across papers challenging. Relatedly, we did not record how many of the reviewed XAI papers (a) included user experiments with functional metrics (XAI completeness, faithfulness, robustness, etc.) tests, (b) introduced new XAI techniques, or (c) offered only comparisons between different techniques in a specific domain. This information could have revealed different correlations regarding the generalizability of study results in subsets of the reviewed papers. Future research on XAI user study generalizations that includes this information would be desirable.

Moreover, we searched only three major databases for XAI research, and focused on English publications, and papers with >10 citations for the snowballing. We may therefore have missed important XAI user studies.

Additionally, our sample contained 14 not yet peer-reviewed arXiv papers. Focusing only on peer-reviewed articles might change results. However, when we re-ran the analyses only with peer-reviewed papers, our key findings retained the same trend (see Appendix). The only difference was that quantitative papers with broader conclusions now also had larger samples ($p = .027$). However, our focus in this study was on examining XAI researchers' extrapolation tendency in general. Analyzing only peer-reviewed papers is less likely to provide insights into researchers' general disposition to extrapolate rather than into what they do if their inferences are peer-reviewed. Since including not yet peer-reviewed papers is more informative on researchers' general disposition and our key trends remained the same, we kept these papers in our analyses.

Finally, we extracted our data from papers manually, not automatically. Human error could have occurred. However, to reduce this risk, we reviewed papers independently, crosschecked each other's classifications, had two author-independent raters classify subsets of the data, calculated inter-rater agreement, and made our data publicly available for reproduction.

## 5. Principles for more inclusive XAI testing

XAI user studies with underpowered samples, missing sample size justifications, and overgeneralizations may lead to an oversight of many people's explanatory needs regarding the AI systems they may encounter. This can hinder the development of ethical AI that uses XAI to implement explainability. We propose three methodological principles to counteract these problems:

(1) *Principle of sample size justification.* Studies of new technologies may frequently start small due to funding or feasibility concerns. This may often be acceptable to explore the usability of models and facilitate developmental progress. However, to promote best scientific practice, AI journals should require user studies to include (where feasible) power analyses, or other sample size justifications. For XAI researchers who struggle to find method-specific sample size rationales, we recommend three papers with overviews of sample size justifications: [4, 7, 8].

(2) *Principle of reporting relevant background.* Since having technical affinity, AI expertise, or a computer science background is known to influence XAI user perception [1], it is important to question participants about it to control for it. This may also apply to other factors (e.g., gender). AI journals should ask XAI study authors to consider, report, and justify the inclusion and exclusion of relevant control variables. Overlooking variables to control can facilitate false positives whereas including too many can facilitate false negatives. For best practice on control variable usage, we recommend [16].

(3) *Principle of generalization checks.* Overgeneralizations may result when researchers need to persuade journals, funding bodies, and policy-makers of their studies' importance, or when writing guidelines require more condensed language. We recommend that AI journals reconsider their common emphasis on condensed formats and adopt a principle for authors to provide generality constraint statements in their papers, i.e., statements that articulate generalizability limits and justify the scope of result-related claims. For guidance on what to include in such statements, we recommend [19].

We also suggest that XAI authors use a checklist containing reminders to (a) mention relevant qualifiers (statistical distributions, percentages, 'U.S. users', etc.) to tailor their conclusions to the evidence, and (b) consider using the past tense when reporting results, which restricts results to subsets of individuals [20]. To illustrate this, in the Appendix (Table II), we present restricted reformulations of the unrestricted claims from Table 1.

## 6. Conclusion

In this systematic analysis of XAI user studies, we found that many XAI researchers did not follow best scientific practice. They did not justify the size of the samples tested, leaving it unclear whether the samples were too small to detect differences and generalize to the target population, or too large, inflating practically irrelevant differences. Yet, most of the studies nonetheless contained broad conclusions extending beyond their samples and study populations to users, people, etc. in general. Since there was no evidence that these broad conclusions were correlated with larger, more representative samples, and researchers had often not checked for a technical background among their participants that may undermine generalizability, the conclusions that we found in most papers were overgeneralizations. Given the important and valuable role that XAI user research plays for informing stake-holders' assessments of whether a given XAI model implements ethical AI frameworks, future XAI user study methods should be improved such that sample size justifications become routine in XAI research and overgeneralizations are avoided.

**Short Author Bios**

**First Author** (study design, data collection, data analysis, argumentation, editing): Uwe Peters is a Postdoctoral Researcher at the Leverhulme Centre for the Future of Intelligence, University of Cambridge, UK, and the Center for Science and Thought, University of Bonn, Germany. His research interests include XAI, algorithmic bias, and scientific methodology. He holds a M.Sc. (Psychology and Neuroscience) and a Ph.D. (Philosophy) from King's College London. Contact: up228@cam.ac.uk

**Second Author** (data collection, editing): Mary Carman is a Lecturer in Philosophy at the University of Witwatersrand, Johannesburg, South Africa. Her research interests include ethical and inclusive AI research and application. Carman received her Ph.D. (Philosophy) from King's College London. She is a co-chair of the working group on 'Digital Earth Governance and Ethics' of the International Society for Digital Earth and an affiliated expert with the Global AI Ethics Institute. Contact: mary.carman@wits.ac.za

# Supplementary Materials

## Search strings (used in July 2022)

**SCOPUS**:

TITLE-ABS-KEY ("XAI" OR "Explainable AI" OR "transparent AI" OR "interpretable AI" OR "accountable AI" OR "AI explainability" OR "AI transparency" OR "AI accountability" OR "AI interpretability" OR "model explainability" OR "explainable artificial intelligence" OR "explainable ML" OR "explainable machine learning" OR "algorithmic explicability" OR "algorithmic explainability") AND ("end user" OR "end-user" OR "audience" OR "consumer" OR "user" OR "user study" OR "user survey" OR "developer") AND (LIMIT-TO (DOCTYPE,"cp") OR LIMIT-TO (DOCTYPE,"ar") OR LIMIT-TO (DOCTYPE,"ch")) AND (LIMIT-TO (PUBYEAR,2022) OR LIMIT-TO (PUBYEAR,2021) OR LIMIT-TO (PUBYEAR,2020) OR LIMIT-TO (PUBYEAR,2019) OR LIMIT-TO (PUBYEAR,2018) OR LIMIT-TO (PUBYEAR,2017) OR LIMIT-TO (PUBYEAR,2016) OR LIMIT-TO (PUBYEAR,2012)) AND (LIMIT-TO (LANGUAGE,"English"))

**Web of Science**:

(ALL=("XAI" OR "Explainable AI" OR "transparent AI" OR "interpretable AI" OR "accountable AI" OR "AI explainability" OR "AI transparency" OR "AI accountability" OR "AI interpretability" OR "model explainability" OR "explainable artificial intelligence" OR "explainable ML" OR "explainable machine learning" OR "algorithmic explicability" OR "algorithmic explainability")) AND ALL=("end user" OR "end-user" OR "audience" OR "consumer" OR "user" OR "user study" OR "user survey" OR "developer") and Article or Proceedings Papers or Early Access or Book Chapters(Document Types) and English (Languages)

Refined by all 'Publication Years' (2012-01-01 to 2022-12-31)

**ArXiv**:

Query: order: -announced_date_first; size: 50; date_range: from 2012-01-01 to 2022-12-31; classification: Computer Science (cs); include_cross_list: True; terms: AND all="XAI" OR "Explainable AI" OR "transparent AI" OR "interpretable AI" OR "accountable AI" OR "AI explainability" OR "AI transparency" OR "AI accountability" OR "AI interpretability" OR "model explainability" OR "explainable artificial intelligence" OR "explainable ML" OR "explainable machine learning" OR "algorithmic explicability" OR "algorithmic explainability"; AND all="end user" OR "end-user" OR "audience" OR "consumer" OR "user" OR "user study" OR "user survey" OR "developer"

**Link to the OSF platform with the material used for the review:**

https://osf.io/vzndw/

| Year | Number |
|---|---|
| 2012 | 0 |
| 2013 | 1 |
| 2015 | 1 |
| 2016 | 2 |
| 2017 | 1 |
| 2018 | 6 |
| 2019 | 19 |
| 2020 | 46 |
| 2021 | 84 |
| July 2022 | 60 |
| **Total** | **220** |

**Table I.** Numbers of XAI user study papers published between January 2012 and July 2022

**Main results with not yet peer-reviewed papers ($n = 14$) excluded:**

(1) Sample size justification: No = 82.5% ($n$ 170) vs. Yes = 17.5% ($n = 36$)

(2) Scope of conclusion: 'restricted' = 30.6% ($n = 63$) vs. 'unrestricted' = 69.4% ($n = 143$)

(3) Asked about technical background: No = 54.4% ($n = 112$) vs. Yes = 45.6% ($n = 94$).

(4) Sample size justification related to sample size: papers with justification ($n = 181$, *Mean rank* = 99.03) vs. papers without justification ($n = 25$, *Mean rank* = 135.88, $U = 1453.00$, $z = –2.898$, $p = .004$); correlation: $r_{rb}$ (204) = .202, 95% CI [.064, .334], $p = .004$.

(5) Broader conclusions (quantitative papers) larger samples: 'unrestricted' papers ($n = 94$, *Mean rank* = 62.42) vs. 'restricted' papers ($n = 23$, *Mean rank* = 45.02, $U = 759.50$, $z = –2.205$, $p = .027$); correlation: $r_{rb}$ (115) = .205, 95% CI [.019, .377], $p = .027$.

| |
|---|
| (1) Unrestricted: "Specifically, our results suggest that users prefer more diverse local explanations when they are presented alone compared to when a global explanation is also available." [12]  <br> Restricted: "Specifically, our results suggest that users **preferred** more diverse local explanations when they **were** presented alone compared to when a global explanation **was** also available." |
| (2) Unrestricted: "We demonstrated that CX-ToM significantly outperforms baselines in improving human understanding of the underlying classification model." [26]  <br> Restricted: "We demonstrated that CX-ToM significantly outperform**ed** baselines in improving **many people's** understanding of the underlying classification model." |
| (3) Unrestricted: "Our pilot study revealed that users are more interested in solutions to errors than they are in just why the error happened." [42]  <br> Restricted: "Our pilot study revealed that **many participants were** more interested in solutions to errors than they are in just why the error happened." |
| (4) Unrestricted: "Explanations lead people to view errors as being 'less incorrect', but they do not improve trust." [130]  <br> Restricted: "Explanations **led participants** to view errors as being 'less incorrect', but they **did** not improve trust." |
| (5) Unrestricted: "People prefer item-centric but not user-centric or socio-centric explanations." [142]  <br> Restricted: "**Participants preferred** item-centric but not user-centric or socio-centric explanations." |
| (6) Unrestricted "Results indicate that human users tend to favor explanations about policy rather than about single actions." [205]  <br> Restricted: "Results indicate that human users tend**ed** to favor explanations about policy rather than about single actions." |
| (7) Unrestricted: "Our findings suggest that people do not fully trust algorithms for various reasons, even when they have a better idea of how the algorithm works." [213]  <br> Restricted: Our findings suggest that **participants did** not fully trust algorithms for various reasons, even when they **had** a better idea of how the algorithm worked." |

**Table II.** Restricted versions of the unrestricted claims from Table 1 (restricting parts in bold)